\documentclass[10pt,twocolumn,letterpaper]{article}
\PassOptionsToPackage{table,dvipsnames}{xcolor}
\usepackage[pagenumbers]{iccv}
\usepackage{multirow}
\usepackage{footmisc}
\definecolor{lightblue}{HTML}{e1f4ff}

\newcommand{\cellcolorlightblue}{\cellcolor{lightblue}}

\definecolor{iccvblue}{rgb}{0.21,0.49,0.74}
\usepackage[pagebackref,breaklinks,colorlinks,allcolors=iccvblue]{hyperref}

\newcommand{\customfootnotetext}[2]{% 
  \begingroup
  \setlength{\skip\footins}{0pt}
  \renewcommand{\thefootnote}{#1}%
  \footnotetext{#2}%
  \endgroup
}

\title{YOLO-Count: Differentiable Object Counting for Text-to-Image Generation}
\author{%
  Guanning Zeng\textsuperscript{1}\textsuperscript{*} \quad
  Xiang Zhang\textsuperscript{2} \quad
  Zirui Wang\textsuperscript{3} \quad
  Haiyang Xu\textsuperscript{2} \\
  Zeyuan Chen\textsuperscript{2} \quad
  Bingnan Li\textsuperscript{2} \quad
  Zhuowen Tu\textsuperscript{2} \\
  \textsuperscript{1}Tsinghua University \quad 
  \textsuperscript{2}UC San Diego \quad
  \textsuperscript{3}UC Berkeley
}

\begin{document}

\twocolumn[{
\renewcommand\twocolumn[1][]{#1}
    \maketitle
    \begin{center}
        \vspace{-15pt}
        \centering
        \includegraphics[width=\linewidth]{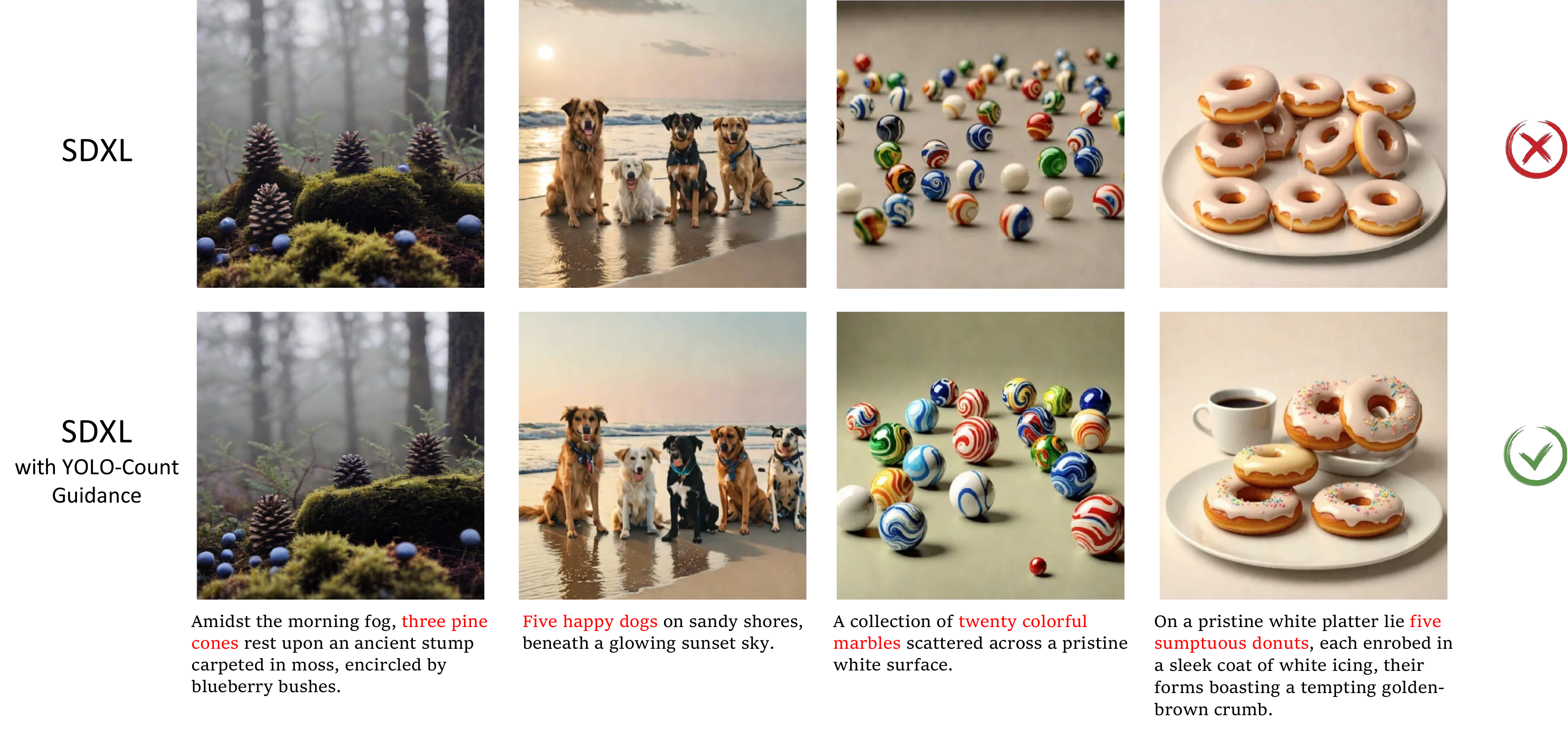}
        \captionsetup[figure]{hypcap=false}
        \captionof{figure}{{\bf Demonstrations of YOLO-Count's object quantity controllability} for text-to-image generation. Incorporating YOLO-Count as a differentiable guidance module over a strong baseline (SDXL~\cite{sdxlturbo2024}) substantially improves alignment between text-specified object quantities and the generated images.}
        \vspace{1em}
    \label{fig:demo}
    \end{center}
}]

\customfootnotetext{*}{Work done during internship at UC San Diego.}
\begin{abstract}
We propose YOLO-Count, a differentiable open-vocabulary object counting model that tackles both general counting challenges and enables precise quantity control for text-to-image (T2I) generation. A core contribution is the `cardinality' map, a novel regression target that accounts for variations in object size and spatial distribution. Leveraging representation alignment and a hybrid strong-weak supervision scheme, YOLO-Count bridges the gap between open-vocabulary counting and T2I generation control. Its fully differentiable architecture facilitates gradient-based optimization, enabling accurate object count estimation and fine-grained guidance for generative models. Extensive experiments demonstrate that YOLO-Count achieves state-of-the-art counting accuracy while providing robust and effective quantity control for T2I systems.
\vspace*{-1cm}
\end{abstract}

\section{Introduction}
Text-to-image (T2I) generative models have achieved remarkable success in producing high-fidelity images from natural language descriptions. However, ensuring precise alignment with textual specifications, particularly regarding object quantity, remains a significant challenge. While prior research has improved adherence to object layout, attributes, and style through conditional training and guidance mechanisms, accurately controlling the number of objects synthesized within an image remains difficult. Unlike localized attributes, object quantity constitutes a global constraint, requiring models to establish numerical correspondence between language tokens and compositional objects. Consequently, conventional conditional training approaches such as ControlNet~\cite{zhang2023adding} are ill-suited for explicit quantity control. Moreover, the stochastic nature of the denoising process in T2I models introduces ambiguity in object differentiation, further complicating count consistency. Recent conditional guidance methods, such as BoxDiff~\cite{xie2023boxdiff} and Ranni~\cite{feng2024ranni}, address aspects of spatial layout, object attributes, and semantic panel conditioning. However, these methods lack a direct and principled mechanism for precise quantity control, leaving a critical gap in bridging linguistic numeracy and visual synthesis.

In this work, we propose YOLO-Count, an open-vocabulary object counting model built on the YOLO architecture. YOLO-Count is a fully differentiable, regression-based model that demonstrates high accuracy, computational efficiency, and open-vocabulary capabilities. A key contribution is the introduction of the \emph{cardinality map}, a novel representation that encodes object quantity while preserving awareness of object size and spatial location. Unlike traditional density maps, which apply Gaussian kernels at object centers, the cardinality map distributes quantity scores across object instances, improving accuracy and robustness to scale variation. Furthermore, YOLO-Count leverages representation alignment and a hybrid strong-weak supervision strategy, enabling the use of large-scale instance segmentation datasets without reliance on computationally expensive pre-trained visual encoders.

Beyond generic object counting, we are motivated to apply YOLO-Count for precise control of object quantities in text-to-image (T2I) generation. This is achieved by employing YOLO-Count as a differentiable guidance module~\cite{bansal2023universal}, where gradient signals from the counting model steer the generative process toward numerical consistency. While prior research has predominantly focused on guidance algorithms for attributes and layout, explicit quantity control remains underexplored. We argue that an ideal object counting model for T2I applications should possess four key properties: (1) \emph{full differentiability} \wrt the input image; (2) \emph{open-vocabulary} capability for diverse object categories; (3) \emph{cross-scale generalization} to varying object sizes; (4) \emph{computational efficiency} for practical deployment.

Constructing such a model introduces several challenges. First, state-of-the-art counting approaches~\cite{pelhan2024dave, amini2025countgd} are often detection-based, producing outputs that preclude gradient propagation. Second, existing counting datasets such as FSC147~\cite{m_Ranjan-etal-CVPR21} or CARPK~\cite{hsieh2017drone} are limited in scale and category diversity, hindering open-vocabulary generalization. Third, while large-scale vision encoders (\eg, CLIP~\cite{radford2021learning} or GroundingDINO~\cite{liu2024grounding, ren2024grounding}) can alleviate data limitations, they impose significant computational overhead.

To address these issues, we integrate YOLO-Count with textual inversion~\cite{DBLP:conf/iclr/GalAAPBCC23, zafar2024iterativeobjectcountoptimization} to achieve precise quantity control in T2I generation. Extensive experiments demonstrate that YOLO-Count achieves state-of-the-art accuracy on counting benchmarks, outperforms density-based and detection-based counting models, and substantially improves object quantity controllability in T2I generation.

Our contributions are summarized as follows:

\begin{itemize}
\item We introduce the \emph{cardinality map}, a novel regression target that improves object counting accuracy compared to density maps.
\item We develop YOLO-Count, an efficient, open-vocabulary, and \emph{fully differentiable} counting model that achieves state-of-the-art performance and enhances quantity control for T2I generation.
\item We propose \emph{hybrid strong-weak supervision} with representation alignment, enabling effective training using large-scale segmentation datasets without reliance on heavy visual encoders.

\end{itemize}

\section{Related Works}

\subsection{Object Counting Models and Datasets}

Object counting models can be broadly classified according to their category scope into fixed-category counting models~\cite{shi2024focus, wang2024weakly, guo2024regressor} and open-vocabulary counting models~\cite{amini2025countgd, pelhan2024dave, Dai_2024_CVPR}. For controlling object quantities in generative tasks, open-vocabulary counting is essential, as it supports arbitrary object categories without retraining. Based on the type of supervision or guidance, counting models can be further divided into text-guided models~\cite{xu2023zero, zhu2024zero}, visual-exemplar-guided models~\cite{m_Ranjan-etal-CVPR21, huang2024point}, multimodal-guided models~\cite{amini2025countgd}, and reference-less models~\cite{hobley2022learning, liu2022countr, wu2024gca}. For T2I integration, a purely text-guided counting model is preferable to ensure compatibility with prompt-driven generation.

From a methodological perspective, counting models are typically divided into detection-based and regression-based approaches. Detection-based models~\cite{hsieh2017drone, amini2025countgd, nguyen2022few} rely on explicit object detection, filtering instances via thresholds and enumerating discrete counts, which inherently produce non-differentiable integer outputs. In contrast, regression-based models~\cite{arteta2014interactive, arteta2016counting, lempitsky2010learning} predict continuous-valued maps such as density maps~\cite{cho1999neural, marana1997estimation} that represent pixel-wise contributions to the final count. This direct differentiability makes regression-based models particularly suitable for gradient-based control in generative pipelines.

Finally, training datasets for object counting are categorized into fixed-category datasets~\cite{hsieh2017drone, gao2020nwpu, idrees2018composition} and open-vocabulary datasets~\cite{m_Ranjan-etal-CVPR21, AminiNaieni23}. Open-vocabulary datasets provide images containing diverse object categories and instance counts, but are expensive to collect and annotate~\cite{m_Ranjan-etal-CVPR21}. For example, the widely used FSC147 dataset includes only 3,659 training images, which limits scale and diversity. To address this, recent works~\cite{jiang2023clip, amini2025countgd} incorporate large-scale pre-trained visual backbones (\eg, CLIP~\cite{radford2021learning} and GroundingDINO~\cite{liu2024grounding}) and fine-tune them on smaller counting datasets to enhance open-vocabulary generalization.

\begin{figure*}[htbp]
    \centering
    \includegraphics[width=\linewidth]{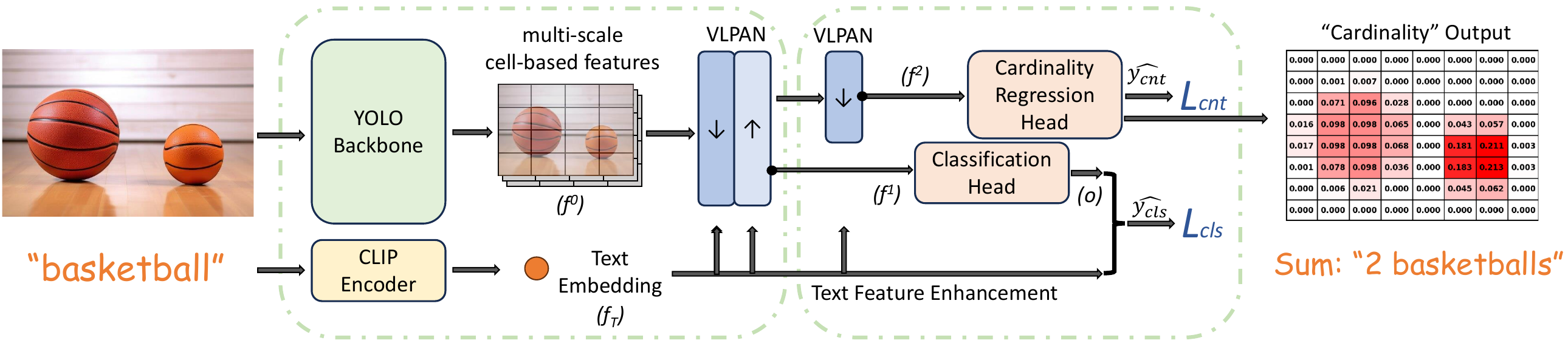}
    \caption{{\bf YOLO-Count Model Overview}. YOLO-Count comprises a YOLO backbone, CLIP text encoder, vision-language path aggregation network (VLPAN), cardinality regression head and classification head. Built upon the YOLO-World~\cite{Cheng2024YOLOWorld} architecture, the cardinality head predicts a cardinality map. The final object quantity is obtained by summing over the cardinality map.}
    \label{fig:pipeline}
\end{figure*}

\subsection{Controllable Text-to-Image Generation}
Controllable text-to-image (T2I) generation methods can be broadly categorized into two paradigms: training-based methods~\cite{zhang2023adding, zhao2023uni, hu2023cocktail} and guidance-based methods~\cite{zhao2023loco, bansal2023universal, yu2023freedom}. Training-based approaches, such as ControlNet~\cite{zhang2023adding}, IP-Adapter~\cite{ye2023ip}, and GLIGEN~\cite{li2023gligen}, inject conditional inputs directly into the generative model through additional network branches or adapters. While effective, these methods rely on large-scale training datasets annotated with the corresponding conditions. In contrast, guidance-based approaches, including BoxDiff~\cite{xie2023boxdiff}, Attend-and-Excite~\cite{chefer2023attend}, and Separate-and-Enhance~\cite{bao2024separate}, control generation by manipulating the diffusion process at inference time, eliminating the need for retraining. Many of these methods exploit the interpretability of cross-attention mechanisms~\cite{liu2024towards} to steer image synthesis. However, cross-attention is primarily effective for distinguishing object categories rather than differentiating multiple instances of the same category. As a result, existing controllable T2I techniques excel at localized attribute binding~\cite{guo2024initno, zhuang2025magnet} and layout control~\cite{zhou2024migc, zheng2023layoutdiffusion}, but struggle with enforcing global constraints such as precise object quantity.

\subsection{Object Quantity Control for T2I Models}
Research on explicit object quantity control in text-to-image (T2I) models remains limited. \cite{kang2023counting} pioneers the use of universal diffusion guidance for quantity control, representing the first attempt to directly address this challenge. \cite{binyamin2024make} introduces an attention-based representation for counting objects, but their approach is constrained to controlling small quantities (ranging from 1 to 10). More recently, prompt-tuning approaches~\cite{zafar2024iterativeobjectcountoptimization, sun2024quota} have been proposed to incorporate numerical cues into the text embedding space, enabling limited quantity control without modifying the underlying diffusion model. However, these methods still struggle with accurate control over larger counts.

\section{Methods}
\subsection{Model Overview}
Our proposed YOLO-Count builds upon the YOLO-World architecture~\cite{Cheng2024YOLOWorld} and consists of three primary components: (1) a vision backbone, (2) a vision-language path aggregation network (VLPAN), and (3) prediction heads. \cref{fig:pipeline} illustrates the overall pipeline and highlights our key architectural modifications.

\paragraph{Vision Backbone.} The vision backbone in YOLO-Count follows the design of YOLOv8l~\cite{yolov8_ultralytics} and YOLO-World-L~\cite{Cheng2024YOLOWorld}. It comprises five stages of convolutional modules (ConvModules) and cross-stage partial layers (CSPLayers). Given an input image $I \in \mathbb{R}^{640 \times 640 \times 3}$, the backbone extracts multiscale visual features at three resolutions:
$$
f^0 =[f_{80\times 80}, f_{40 \times 40}, f_{20 \times 20}] = \mathrm{VisualBackbone}(I)
$$

\paragraph{Vision-Language Path Aggregation Network (VLPAN).}  The VLPAN is designed to fuse visual features with textual semantics and aggregate information across scales. Inheriting from YOLO-World, it employs both top-down and bottom-up pathways, but with key enhancements: (1) T-CSPLayers: standard CSPLayers are replaced by T-CSPLayers, which integrate sigmoid attention blocks to modulate visual features based on precomputed CLIP text embeddings~\cite{radford2021learning}. (2) Extended Top-Down Fusion: to better preserve fine-grained spatial details, an additional top-down pathway is introduced following the initial bidirectional aggregation, maximizing high-resolution feature utilization, which is critical for accurate counting regression. The enhanced VLPAN is formulated as:
$$
[f^1, f^2] = \mathrm{VLPAN}(f^0, f_\mathrm{T})
$$
where $f_\mathrm{T}$ denotes the CLIP text embedding of the category, $f^1$ and $f^2$ represent multimodal features for classification and counting regression, respectively.

\paragraph{Prediction Heads.} Following the VLPAN, several ConvModules are applied to text-aware visual features to aggregate multi-scale signals into a unified $80 \times 80$ resolution. The prediction stage then produces two parallel outputs: (1) a cardinality regression head, which predicts a dense cardinality map for differentiable counting, and  
(2) a classification head, trained with contrastive supervision to ensure robust open-vocabulary capability. These two outputs jointly enable YOLO-Count to provide accurate, differentiable count estimates while maintaining strong category generalization, as shown on the right side of \cref{fig:pipeline}.
$$
\left\{
\begin{aligned}
o_\mathrm{cls} &= \mathrm{ClassificationHead}(f^1) \\
\hat{y}_\mathrm{cnt} &= \mathrm{CountingHead}(f^2)
\end{aligned}
\right.
$$

\subsection{Cardinality Map Regression}
We introduce the concept of the \emph{cardinality map}, a novel regression target designed to address the inherent ambiguities of density-based counting.

Density-based counting models formulate the counting loss as:
\begin{equation}
\mathcal{L}_\mathrm{cnt} = \left| \hat{y}_\mathrm{cnt} - y_\mathrm{den} \right|
\end{equation}
where $y_\mathrm{den}$ is the density map. For an image containing $Q$ objects, $y_\mathrm{den}$ is constructed by placing $Q$ Gaussian kernels centered at object locations, with their sum equal to $Q$. While regressing to $y_{\mathrm{den}}$ enables quantity prediction, this representation suffers from two key ambiguities. First, the Gaussian kernel’s center can be placed anywhere within an object’s extent. Second, the kernel radius is arbitrarily chosen and lacks physical meaning. These issues degrade model accuracy for objects with diverse sizes and shapes, as density maps fail to provide a consistent, unambiguous representation.

To overcome these limitations, we replace $y_{\mathrm{den}}$ with a \emph{cardinality map} $y_{\mathrm{car}}$, defined using object masks. Given a binary mask $M_i$ for the $i$-th object instance, with area $N_i = |M_i|$, we uniformly distribute a value of 1 across all pixels within the object and sum contributions across all $K$ objects:
\begin{equation}
y_\mathrm{pixel\;cardinality} = \sum_{i=1}^K \frac{1}{N_i} M_i
\end{equation}

We then downsample this pixel-level cardinality map to a grid-based representation by summing within each grid cell:
\begin{equation}
    y_{\mathrm{car}}(u, v) = \sum_{(i,j) \in \Omega_{u,v}} y_{\mathrm{pixel\;cardinality}}(i,j)
    \label{eq:cardinality_map}
\end{equation}
where $\Omega_{u,v}$ is the set of pixel coordinates within grid cell $(u,v)$. By construction, the total sum of the cardinality map equals the true object count:
\begin{equation}
    \sum_{u,v} y_{\mathrm{car}}(u,v) = Q
\end{equation}

Unlike density maps, which concentrate mass at object centers and often ignore large portions of extended objects, the cardinality map uniformly covers the entire spatial extent of each object. This yields a unique, unambiguous representation that is robust to variations in object size and shape, making it better suited for differentiable regression-based counting.

\subsection{Representation Alignment}\label{section:rep-align}

Here we describe our method using the single-category counting scenario for simplicity, where the model is designed to count instances of a specific category specified by the user. To achieve this, we adapt the contrastive learning framework into a binary classification task, where each pixel is classified as either belonging to the target category or not. This additional branch aligns the visual and textual representations during training, ensuring the model effectively localizes instances of the specified category. Specifically, the classification loss is formulated as:
\begin{equation}
\mathcal{L}_{\mathrm{cls}} = \mathrm{BCELoss}(\hat{y}_{\mathrm{cls}}, y_{\mathrm{cls}}) \label{eqn:classification}
\end{equation}
where $y_{\mathrm{cls}}{(i,j)} \in \{0,1\}$ is the binary ground truth label indicating whether pixel $(i,j)$ belongs to the target category, and $\hat{y}_{\mathrm{cls}}{(i,j)} \in [0,1]$ is the predicted probability obtained by projecting the visual feature $o_{\mathrm{cls}}$ and the text embedding $f_\mathrm{T}$ into a shared multimodal space and applying a sigmoid activation to their inner product, similar to SigLIP~\cite{zhai2023siglip}.

\subsection{Hybrid Strong-Weak Training}\label{section:hybrid-training}
Training an object-counting model typically necessitates a specialized counting dataset, where each image contains multiple instances of a single category. Regression-based counting models rely on corresponding density maps, which are costly to produce and task-specific. To overcome this data limitation, we propose a hybrid strong-weak training technique that enables cardinality regression using both instance segmentation datasets and counting datasets. This approach comprises two stages: strong-supervision pretraining and weak-supervision finetuning.

\begin{figure}[htbp]
    \centering
    \includegraphics[width=\linewidth]{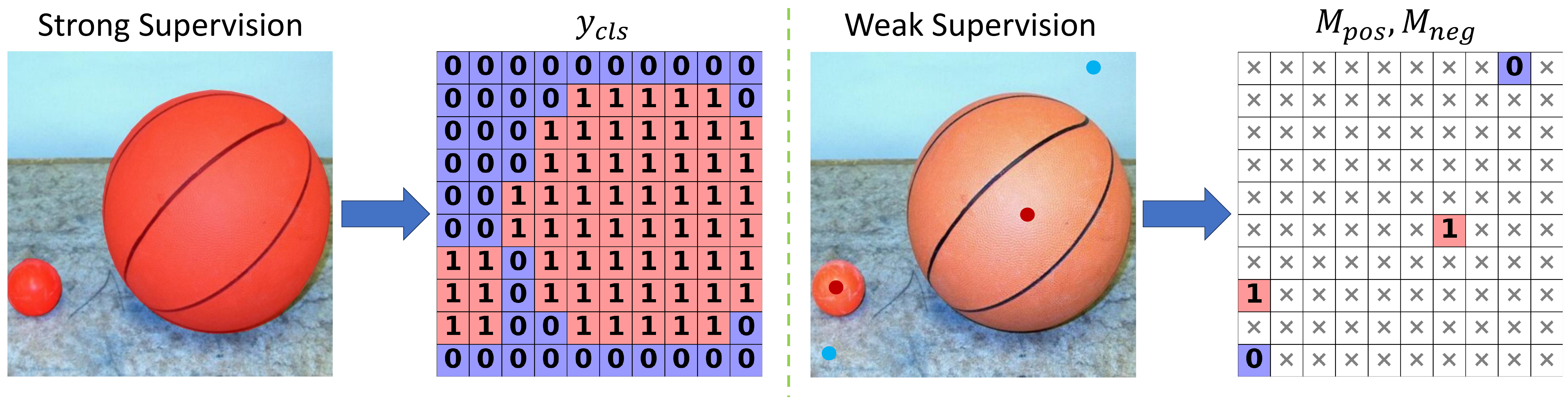}
    \caption{{\bf Illustration of hybrid strong-weak supervision.} Instance segmentation masks provide dense, per-grid category labels, while counting annotations offer sparse category labels limited to annotated points. By combining these two forms of supervision, we enable effective training of the object counting model using both precise (strong) and sparse (weak) annotations.}
    \label{fig:strong_weak}
\end{figure}

\subsubsection{Strong-Supervision Pretraining}
We first pretrain the model on instance segmentation datasets where precise per-instance masks are available. We construct cardinality maps $y_{\mathrm{car}}$ (as defined in \cref{eq:cardinality_map}) and binary classification masks $y_{\mathrm{cls}} \in \{0,1\}^{H \times W}$. The pretraining objectives are as follows:
\begin{equation}
\mathcal{L}_{\mathrm{cnt}}^{\mathrm{strong}} = \left| \hat{y}_{\mathrm{cnt}} - y_{\mathrm{car}} \right|
\end{equation}
\begin{equation}
\mathcal{L}_{\mathrm{total}}^{\mathrm{strong}} = \alpha_1 \mathcal{L}_{\mathrm{cnt}}^{\mathrm{strong}} + \beta_1 \mathcal{L}_{\mathrm{cls}}^{\mathrm{strong}}
\end{equation}
where $\alpha_1$ and $\beta_1$ are weighting coefficients. $\mathcal{L}_{\mathrm{cls}}^{\mathrm{strong}}$ is the same $\mathcal{L}_\mathrm{cls}$ defined in \cref{eqn:classification}. This stage provides precise pixel-level supervision for both cardinality regression and category-specific classification, establishing a strong model initialization.

\subsubsection{Weak-Supervision Finetuning}  
To better adapt YOLO-Count to dense counting scenarios, we perform weak-supervision finetuning on counting datasets that provide sparse point-level annotations. Each image is annotated with instance points $\mathcal{P} = \{(x_i, y_i)\}_{i=1}^K$, where $K$ is both the total count of objects in the image and the number of annotated points. Weak supervision comprises the following two components:

\noindent \textbf{(1) Sparse Classification Labels:} Positive labels are derived from annotated points, forming $M_{\mathrm{pos}} \in \{0,1\}^{H \times W}$ where $M_{\mathrm{pos}}(i,j)=1$ if $(i,j) \in \mathcal{P}$. Negative labels $M_{\mathrm{neg}}$ are sampled from background regions. 

\noindent \textbf{(2) Total Count Consistency:} The predicted total count must match the ground truth count $K$.  

The weak supervision losses are then formulated as follows:
\begin{equation}
\begin{aligned}
\mathcal{L}_{\mathrm{cls}}^{\mathrm{weak}} = - \frac{1}{|\Omega|} \sum_{p \in \Omega} 
\big[ & M_{\mathrm{pos}}(p) \log \hat{y}_{\mathrm{cls}}(p) + \\
      & M_{\mathrm{neg}}(p) \log \left(1 - \hat{y}_{\mathrm{cls}}(p) \right) \big]
\end{aligned}
\end{equation}
\begin{equation}
\mathcal{L}_{\mathrm{cnt}}^{\mathrm{weak}} = \left| \left( \sum_{p} \hat{y}_{\mathrm{cnt}}{(p)} \right) - K \right|
\end{equation}
\begin{equation}
\mathcal{L}_{\mathrm{total}}^{\mathrm{weak}} = \alpha_2 \mathcal{L}_{\mathrm{cnt}}^{\mathrm{weak}} + \beta_2 \mathcal{L}_{\mathrm{cls}}^{\mathrm{weak}},
\end{equation}
where $\Omega = \{ (i,j) \, | \, M_\mathrm{pos}(i,j)=1 \lor M_\mathrm{neg}(i,j)=1 \}$ denotes the annotated pixel locations (see \cref{fig:strong_weak}).

This hybrid training scheme effectively leverages large-scale instance segmentation datasets for pretraining and adapts to limited counting datasets during finetuning, enabling robust and data-efficient model training.

\subsection{Counting-Controlled Generation}\label{section:generation}
Following \cite{zafar2024iterativeobjectcountoptimization}, we adopt a textual inversion-based approach for counting-controlled generation. The text-to-image (T2I) model first synthesizes an initial image from a text prompt, which may not accurately reflect the desired object quantity. We then feed the generated image and its target category into the YOLO-Count model to estimate the predicted quantity and compute a guidance loss based on the deviation from the required count $Q_{\mathrm{req}}$:
\begin{equation}
\mathcal{L}_{\mathrm{guide}} = \left|\left(\sum_p \hat{y}_\mathrm{cnt}(p) \right) - Q_\mathrm{req}\right|
\label{guidance-loss}
\end{equation}

We iteratively update a learnable counting token inserted into the text sequence via the gradient of $\mathcal{L}_{\mathrm{guide}}$. This process continues until convergence, effectively steering the T2I model toward images with the desired object quantity.

\section{Experiments}
\subsection{Setups}
\subsubsection{Training Datasets}

\noindent \textbf{FSC147}~\cite{m_Ranjan-etal-CVPR21} is an object counting dataset comprising 6,135 images across 89 training classes, 29 validation classes, and 29 test classes, with no overlap between the splits. Following \cite{amini2025countgd}, we applied corrections to several images containing erroneous annotations to ensure label consistency.

\noindent \textbf{LVIS v1.0}~\cite{gupta2019lvis} is a large-scale, long-tailed dataset containing 1,203 object categories, sharing images with MSCOCO~\cite{lin2014microsoft}. Originally designed for instance segmentation with a large vocabulary, we employ its validation set to evaluate counting accuracy across diverse and open-vocabulary categories.

\subsubsection{Evaluation Metrics}
Following prior work~\cite{m_Ranjan-etal-CVPR21, hsieh2017drone, idrees2018composition}, we employ mean absolute error (MAE) and root mean square error (RMSE) to evaluate counting performance. MAE measures the average absolute deviation between predicted and ground-truth counts, while RMSE penalizes larger errors more heavily, capturing both accuracy and precision.

\begin{table*}[htbp]
\centering
\resizebox{\linewidth}{!}{
\begin{tabular}{lrrcrrcrrcrrcrr}
\toprule
 & \multicolumn{2}{c}{FSC-Test~\cite{m_Ranjan-etal-CVPR21}} & & \multicolumn{2}{c}{FSC-Val~\cite{m_Ranjan-etal-CVPR21}} & & \multicolumn{2}{c}{LVIS~\cite{gupta2019lvis}} & & \multicolumn{2}{c}{OImg7-New~\cite{OpenImages}} & & \multicolumn{2}{c}{Obj365-New~\cite{shao2019objects365}} \\
 & \multicolumn{2}{c}{\scriptsize\textit{\textbf{(Seen Categories)}}} & & \multicolumn{2}{c}{\scriptsize\textit{\textbf{(Seen Categories)}}} & & \multicolumn{2}{c}{\scriptsize\textit{\textbf{(Seen for Ours)}}} & & \multicolumn{2}{c}{\scriptsize\textit{\textbf{(Unseen Categories)}}} & & \multicolumn{2}{c}{\scriptsize\textit{\textbf{(Unseen Categories)}}} \\
\cmidrule{2-3} \cmidrule{5-6} \cmidrule{8-9} \cmidrule{11-12} \cmidrule{14-15}
\multirow{-3}{*}[2pt]{Model} & {\small MAE}$\downarrow$ & {\small RMSE}  $\downarrow$ & & {\small MAE}$\downarrow$ & {\small RMSE}$\downarrow$ & & {\small MAE}$\downarrow$ & {\small RMSE}$\downarrow$ & & {\small MAE}$\downarrow$ & {\small RMSE}$\downarrow$ & & {\small MAE}$\downarrow$ & {\small RMSE}$\downarrow$ \\
\midrule
\small\textbf{\emph{Non-Differentiable}} & & & & & & & & & & & & & & \\
CountGD~\cite{amini2025countgd}  & 12.98 & 98.35 & & 12.14 & 47.51 & & 4.84 & 12.45 & & 6.09 & 29.92 & & 3.53 & 10.61 \\
DAVE~\cite{pelhan2024dave}  & 14.90 & 103.42 & & 15.48 & 52.57 & & 5.29 & 11.48 & & 5.31 & 14.24 & & 4.89 & 13.22 \\
\midrule
\small\textbf{\emph{Differentiable}} & & & & & & & & & & & & & & \\
CLIP-Count~\cite{jiang2023clip}  & 17.78 & 106.62 & & 18.79 & 61.18 & & 10.81 & 22.61 & & 14.01 & 30.16 & & 15.48 & 30.28 \\
VLCounter~\cite{kang2024vlcounter}  & 17.05 & 106.16 & & 18.06 & 65.13 & & 8.94 & 23.04 & & 15.32 & 34.41 & & 18.08 & 33.02 \\
CounTX~\cite{AminiNaieni23}  & 15.88 & 106.29 & & 17.10 & 65.61 & & 13.70 & 40.94 & & 17.94 & 48.73 & & 18.76 & 44.80 \\
\cellcolorlightblue YOLO-Count \textbf{(Ours)} & \cellcolorlightblue \textbf{14.80} & \cellcolorlightblue \textbf{96.14} & \cellcolorlightblue & \cellcolorlightblue \textbf{15.43} & \cellcolorlightblue \textbf{58.36} & \cellcolorlightblue & \cellcolorlightblue \textbf{1.65} & \cellcolorlightblue \textbf{6.08} & \cellcolorlightblue & \cellcolorlightblue \textbf{3.72} & \cellcolorlightblue \textbf{11.96} & \cellcolorlightblue & \cellcolorlightblue \textbf{3.28} & \cellcolorlightblue \textbf{9.15} \\
\bottomrule
\end{tabular}
}
\caption{{\bf Comparison of counting accuracy} with existing text-guided object counting models.}\label{cnt_acc}
\vspace{-3mm}
\end{table*}

\subsubsection{Benchmarks of Object Counting}
In addition to FSC147 and LVIS, we introduce two new benchmarks to evaluate open-vocabulary object counting accuracy.

OpenImages V7~\cite{OpenImages} and Objects365~\cite{shao2019objects365} are large-scale open-vocabulary object detection datasets, containing 9 million images across 600 categories and 2 million images across 365 categories, respectively. Based on these datasets, we construct two novel counting benchmarks: \textbf{OpenImg7-New} and \textbf{Obj365-New}, specifically designed to evaluate generalization to unseen categories.  

To ensure category novelty, we first compute CLIP~\cite{radford2021learning} embeddings for all text labels in both datasets. We then filter out categories whose maximum cosine similarity with any LVIS label embedding exceeds 0.7. This procedure yields 47 novel categories from OpenImages V7 and 51 from Objects365. Finally, we retain only images containing these filtered categories, resulting in benchmarks with 14,699 images (OpenImg7-New) and 22,724 images (Obj365-New).  These benchmarks provide a challenging evaluation protocol for assessing the open-vocabulary generalization ability of object counting models beyond the categories seen during FSC147 and LVIS training.

\subsubsection{Benchmarks of Controllable Generation}

To evaluate the precision of quantity-controlled text-to-image (T2I) generation, we construct two benchmarks: \textbf{LargeGen} and \textbf{LargeGen-New}. LargeGen is derived from the FSC147 dataset by selecting the 10 text categories with the highest average object counts per image, providing a benchmark for evaluating generation performance on seen categories. LargeGen-New is built from Obj365-New and OpenImg7-New, thereby assessing generation accuracy on novel categories. These benchmarks enable a systematic evaluation of quantity-guided generation, measuring both seen-category and novel-category performance when using object counting models to steer T2I generation.

\subsubsection{Training YOLO-Count Model}
For YOLO-Count, we first perform a 250-epoch strong-supervision pretraining on the LVIS dataset. The vision backbone is initialized with YOLOv8l weights, while all other modules are randomly initialized. We optimize a composite loss with coefficients $\alpha_1 = 1.0$ for cardinality regression and $\beta_1 = 0.1$ for category classification. The CLIP text encoder remains frozen during training, while all other parameters are updated with different learning rates: $5 \times 10^{-9}$ for the backbone (to preserve pre-trained visual representations) and $1 \times 10^{-5}$ for newly initialized modules. This learning rate strategy stabilizes training and facilitates adaptation of the YOLO backbone to the counting task.

Subsequently, we perform weak-supervised finetuning on FSC147, leveraging the dataset-provided positive point annotations and manually annotating negative labels. For negative labels, we dot approximately 10 background points per image in the FSC147 training set. This annotation process is highly efficient, facilitated through a labeling interface, and requires only about 5 seconds per image on average. During finetuning, we retain a proportion $\gamma$ of LVIS data within each training batch to provide strong supervision and preserve the open-vocabulary capability of YOLO-Count. The model is trained for up to 500 epochs with early stopping based on the mean absolute error (MAE) on the FSC147 validation set.

\subsubsection{Counting Control with Token Optimization}
We integrate YOLO-Count with SDXL-Turbo~\cite{sdxlturbo2024} using a differentiable token optimization strategy for quantity-controlled generation. The pipeline employs single-step inference to balance generation quality and computational efficiency. For each generation task, we iteratively optimize a count token embedding for up to 150 steps using a learning rate of $5\times 10^{-3}$, with early stopping applied if the guidance loss (\cref{guidance-loss}) plateaus for 20 consecutive steps. This process allows direct gradient-based feedback from YOLO-Count to refine the token embedding, effectively steering the T2I model toward producing images with the desired object quantity.

\begin{figure}[b]
    \centering
    \includegraphics[width=\linewidth]{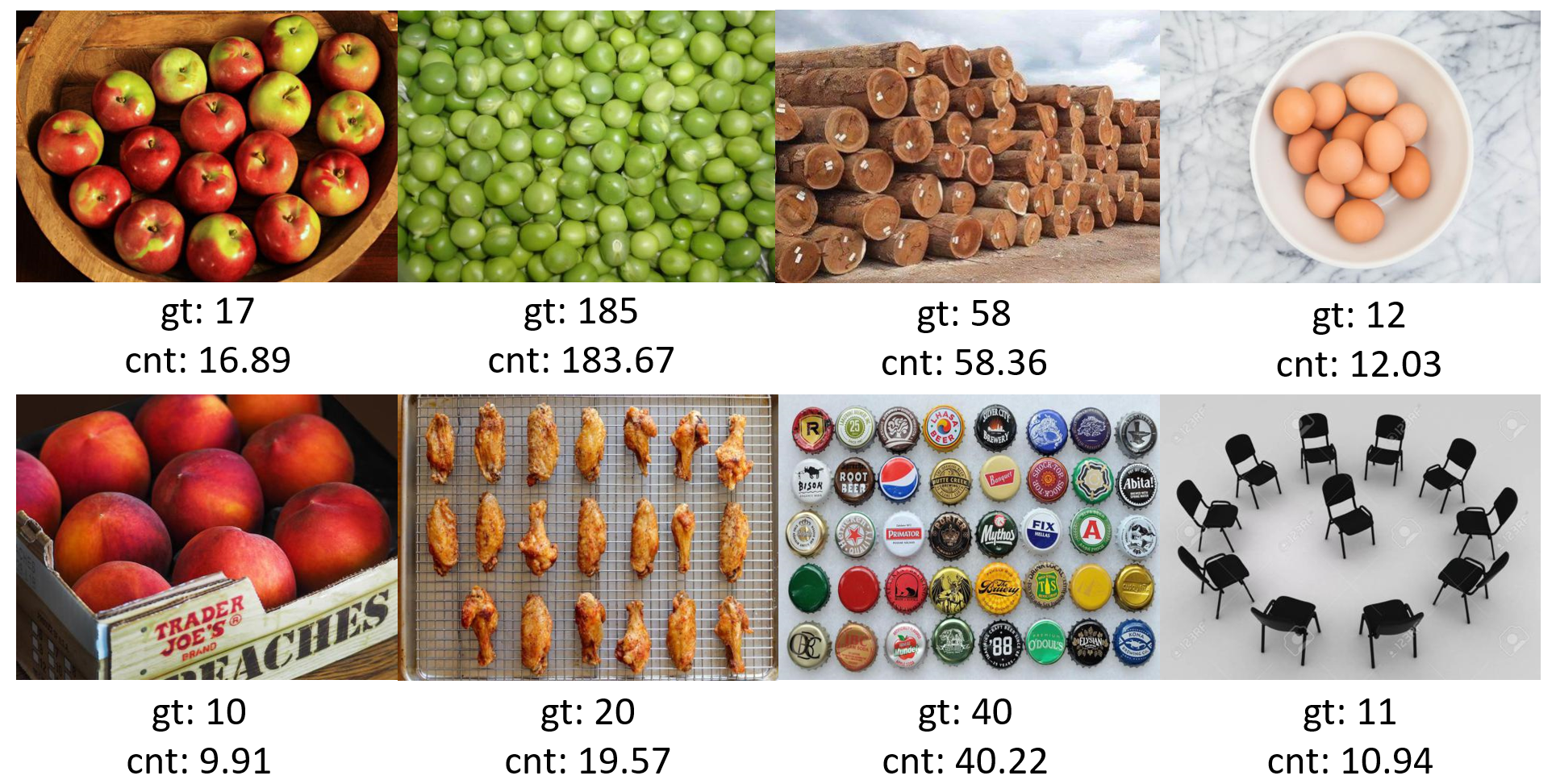}
    \caption{Counting results of YOLO-Count.}
    \label{fig:count_results}
\end{figure}

\subsection{Results of Object Counting}
We evaluate counting accuracy on the FSC147 validation and test sets, a widely used benchmark for object-counting models. For FSC147, we set $\alpha_1 = 0$, $\beta_1 = 1$, $\alpha_2 = 1$, $\beta_2 = 0.1$, and $\gamma = 0.05$ during finetuning. Following prior work such as CountGD~\cite{amini2025countgd} and CLIP-Count~\cite{jiang2023clip}, we adopt an automatic clipping and aggregation strategy to handle images containing large object counts. As shown in \cref{cnt_acc}, YOLO-Count achieves state-of-the-art performance among regression-based counting models. Our MAE and RMSE scores on FSC147 are competitive and closely approach those of CountGD, the current leading non-differentiable counting model. Visual examples in \cref{fig:count_results} further demonstrate that YOLO-Count produces accurate object counts.

We further evaluate the open-vocabulary capability of YOLO-Count by testing its counting accuracy on unseen categories. For LVIS, OpenImg7-New, and Objects365-New, we set $\alpha_1 = \beta_1 = \alpha_2 = \beta_2 = 1$ and $\gamma = 0.5$ for finetuning. The results, shown in \cref{cnt_acc}, demonstrate that exposure to more diverse training data significantly improves YOLO-Count’s ability to generalize to novel categories. Notably, despite having the fewest parameters among the compared counting models, YOLO-Count achieves superior accuracy in open-vocabulary settings.

In summary, \cref{fig:demo,fig:count_results,fig:guide_results,fig:comp_g} illustrate results obtained using the FSC147 checkpoint settings of \cref{cnt_acc}. \cref{fig:comp_reg,fig:bias} present results based on the checkpoint settings for LVIS, OpenImg7, and Objects365 of \cref{cnt_acc}.

\subsection{Results of T2I Quantity Control}
We evaluate the effectiveness of YOLO-Count in guiding T2I models for accurate quantity control. We compare against two series of baselines: (1) extension of previous guidance-based methods~\cite{kang2023counting, binyamin2024make} to large quantities, and (2) the token optimization framework of~\cite{zafar2024iterativeobjectcountoptimization} guided by alternative counting models, including CLIP-Count and CountGD.

For each category in LargeGen and LargeGen-New, we generate 10 images per target quantity, with counts set to 25, 50, 75, and 100. Notably, since CountGD is non-differentiable, we cannot apply the gradient-based loss in \cref{guidance-loss}. Instead, we employ a surrogate cross-entropy objective that encourages a high probability for the target count $Q_{\mathrm{req}}$ while suppressing probabilities for other counts.

\begin{figure}[hb]
    \centering
    \includegraphics[width=\linewidth]{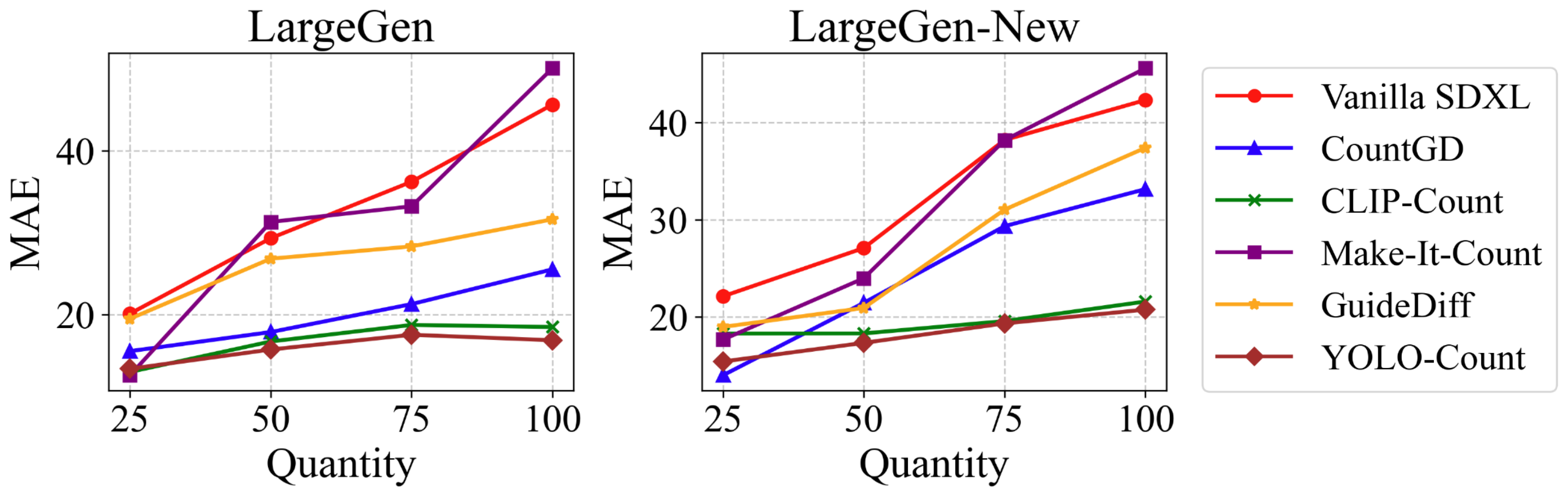}
    \caption{{\bf Quantitive results for T2I quantity control.} Our method significantly reduces the discrepancy between the requested and generated object quantities compared to previous approaches, both on seen and unseen categories.} 
    \label{fig:guide_results}
\end{figure}

We manually evaluate the generated images by counting the objects and comparing them against the target $Q_{\mathrm{req}}$. As shown in \cref{fig:guide_results}, YOLO-Count substantially improves generation accuracy compared to all baselines.

\begin{figure}[ht]
    \centering
    \includegraphics[width=0.9\linewidth]{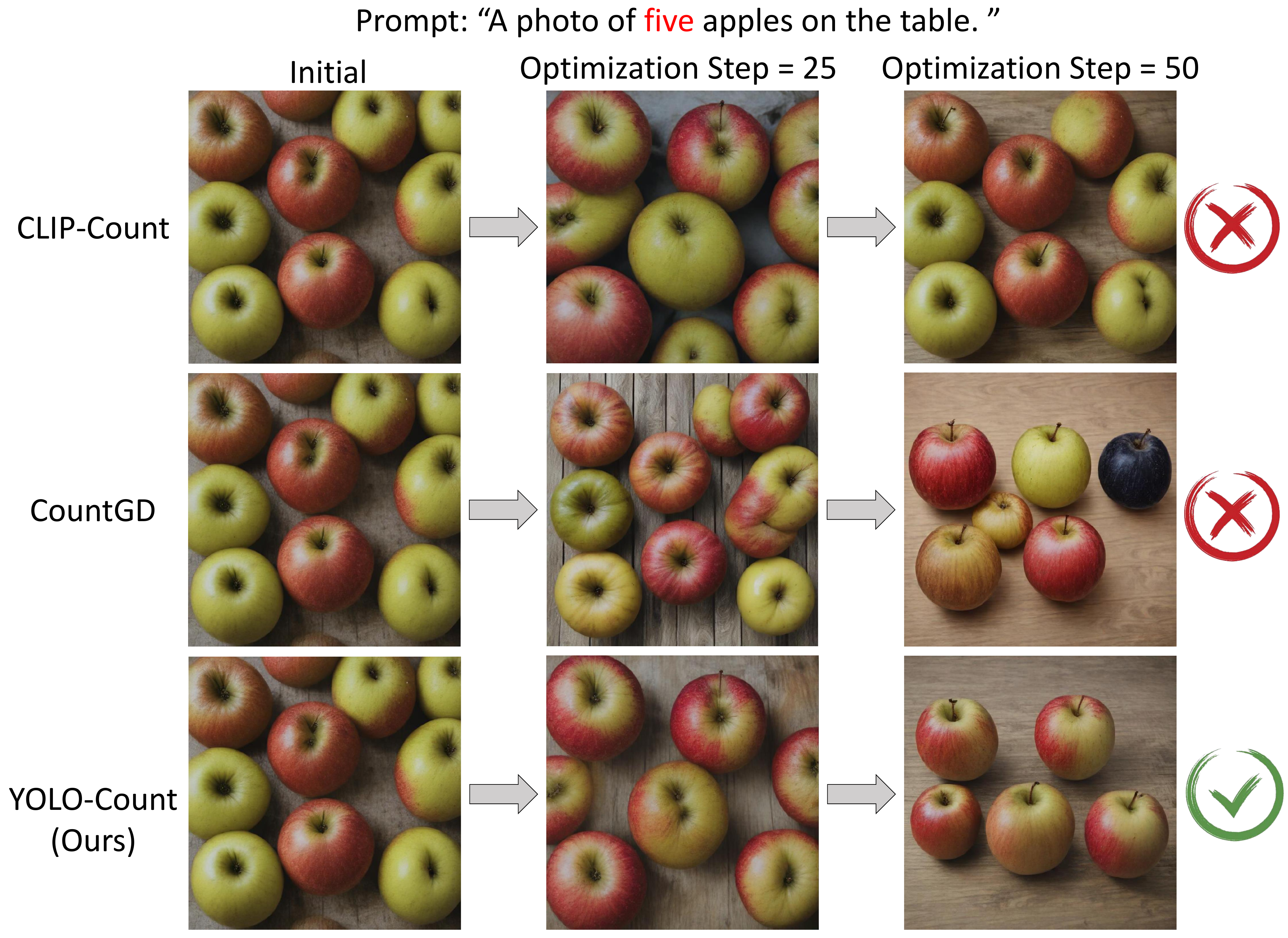}
    \caption{{\bf Comparison of counting-controlled generation} by different models.} 
    \label{fig:comp_g}
\end{figure}

Furthermore, \cref{fig:comp_g} illustrates qualitative differences when generating scenes with large object counts. Under CountGD's surrogate loss $\mathcal{L}_{\mathrm{guide}}^{\mathrm{det}}$, the T2I model fails to adjust object counts and suffers from degraded visual quality. Similarly, density-based counting guidance results in large deviations from $Q_{\mathrm{req}}$, likely due to the domain gap between real counting datasets and synthetic T2I images. In contrast, YOLO-Count provides precise, differentiable guidance signals, enabling accurate quantity control while generalizing effectively to novel categories.
Additional qualitative examples are provided in the \textit{Supplementary Material}.

\subsection{Ablation Study}
We conduct an ablation study to evaluate the contribution of each key component in YOLO-Count. Specifically, we examine the impact of our cardinality map regression, representation alignment, hybrid strong-weak training strategy, and architectural modifications on the performance of regression-based object counting models.

\begin{table}[htbp]
\centering
\resizebox{\linewidth}{!}{
\begin{tabular}{lrrcrr}
\toprule
\multirow{2}{*}[-2pt]{Model} & \multicolumn{2}{c}{FSC-Test \cite{m_Ranjan-etal-CVPR21}} & & \multicolumn{2}{c}{FSC-Val \cite{m_Ranjan-etal-CVPR21}} \\
\cmidrule{2-3} \cmidrule{5-6}
& {\small MAE}$\downarrow$ & {\small RMSE}$\downarrow$ & & {\small MAE}$\downarrow$ & {\small RMSE}$\downarrow$ \\
\midrule
\emph{\textbf{Baseline}} & & & & & \\
YOLO-Count & 14.80 & 96.14 & & 15.43 & 58.36 \\
\midrule
\emph{\textbf{...without}} & & & & & \\
Pretraining & 18.42 & 111.45 & & 19.50 & 88.64 \\
Weak-Supervision & 43.91 & 150.40 & & 43.86 & 124.33  \\
Cardinality Map & 16.71 & 107.24 & & 17.87 & 76.42  \\
Alignment & 17.01 & 110.41 & & 17.57 & 85.54  \\
Additional VLPAN & 16.54 & 106.32 & & 16.89 & 84.40 \\
\bottomrule
\end{tabular}
}
\caption{{\bf Ablation Studies.} w/o Pretraining: no LVIS data; w/o Weak-Supervision: no FSC147 data; w/o Cardinality Map: training on density maps directly; w/o Alignment: no classification branch; w/o Additional VLPAN: both YOLO-Count heads receive same VLPAN feature output.}\label{ablation}
\end{table}

We begin by evaluating the contribution of each stage in the training pipeline. Specifically, we analyze counting accuracy under two conditions: (1) without strong-supervised pretraining and (2) without weak-supervised finetuning. In the first scenario (w/o pretraining), YOLO-Count is trained directly on FSC147 using density map supervision. In the second scenario (w/o finetuning), YOLO-Count is trained solely on LVIS with strong labels, without subsequent finetuning. As shown in \cref{ablation}, the absence of finetuning leads to significantly higher MAE and RMSE, highlighting the domain gap between LVIS and FSC147 in terms of object count distributions. Importantly, combining strong-supervised pretraining on LVIS with weak-supervised finetuning on FSC147 achieves substantial improvements over either stage alone, validating our hybrid training strategy.

\begin{figure}[htbp]
    \centering
    \includegraphics[width=\linewidth]{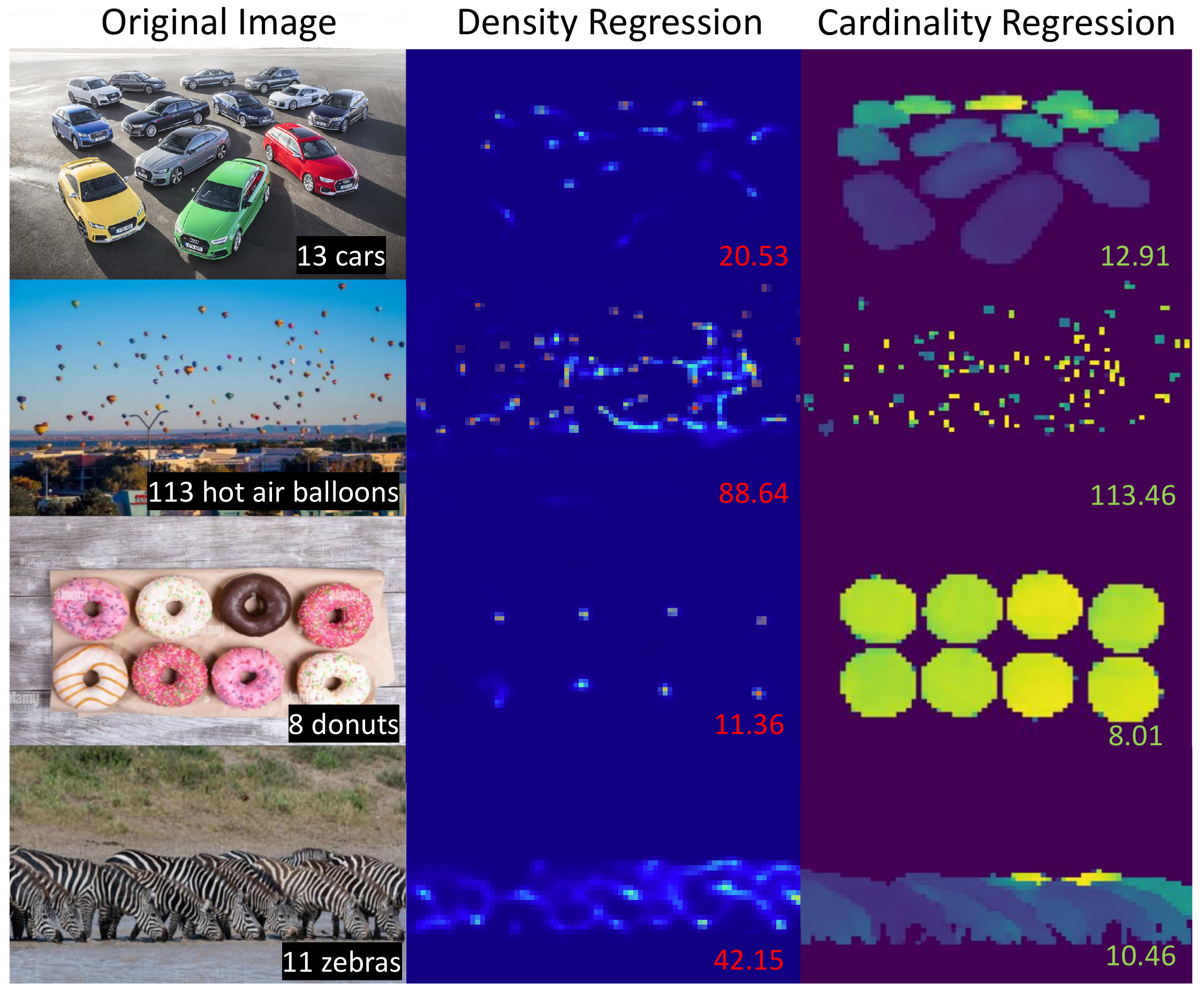}
    \caption{{\bf Comparison of density \vs cardinality regression}.}
    \label{fig:comp_reg}
\end{figure}

Next, we examine the impact of removing the cardinality regression component. In this variant (w/o cardinality), we pretrain on LVIS as in the default setting, but replace cardinality map regression with density map regression during FSC147 training. As observed in \cref{ablation}, the inclusion of cardinality regression yields consistently lower MAE and RMSE compared to density-based regression. Furthermore, visual results in \cref{fig:comp_reg} reveal that density-based regression often suffers from overlapping kernels within a single instance and exhibits a bias toward over-counting in images containing larger-sized objects. This demonstrates that cardinality regression mitigates ambiguities inherent in density maps and improves robustness across varying object scales.

\subsection{Analysis on Size Bias of Counting Models}

\begin{figure}[htbp]
    \centering
    \includegraphics[width=\linewidth]{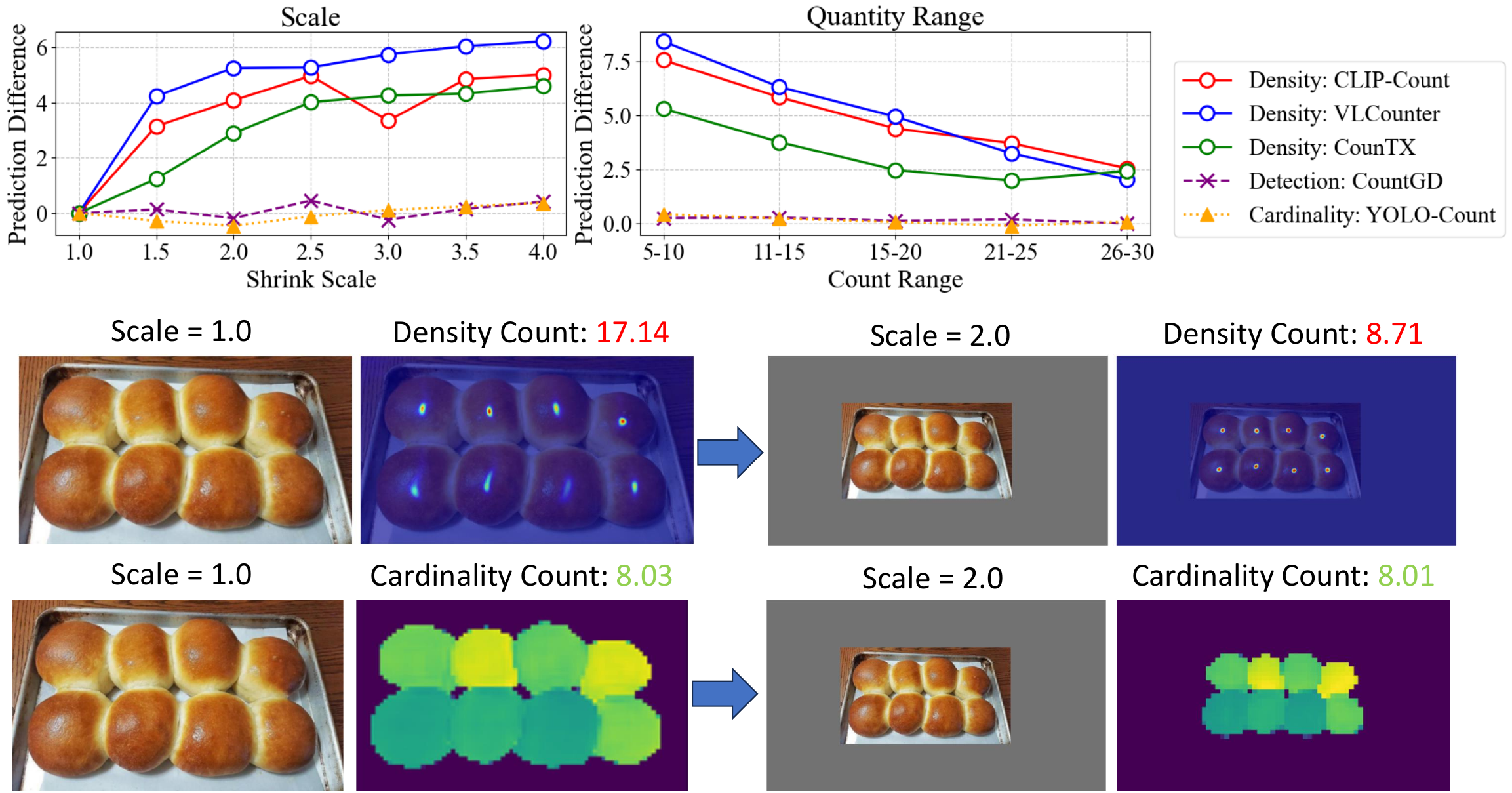}
    \caption{{\bf Comparison of counting bias} for detection, density-based regression and cardinality-based regression models in different image downscaling scales and number of objects.}
    \label{fig:bias}
\end{figure}

We design an experiment to investigate the size bias exhibited by density-based regression methods when handling objects of varying scales. Specifically, we select images from the FSC147 validation and test sets containing no more than 30 objects. Each image is progressively downscaled with a scaling ratio ranging from $1.0$ (original size) to $4.0$, and then padded to restore the original image dimensions. While this process reduces the object sizes, the ground-truth counts remain unchanged. 

The processed images are then fed into several object-counting models: three density-based models (VLCounter~\cite{kang2024vlcounter}, CLIP-Count~\cite{jiang2023clip}, and CounTX~\cite{AminiNaieni23}), one detection-based model (CountGD~\cite{liu2022countr}), and our proposed YOLO-Count model. For each model, we record the predicted counts and analyze their differences relative to predictions on the original (unscaled) images, providing a measure of counting stability under size variation.

As illustrated in \cref{fig:bias}, density-based regression models consistently over-count as object sizes increase, with this bias becoming more pronounced for images containing larger objects. In contrast, YOLO-Count exhibits stability similar to detection-based models, maintaining accurate counts across different object scales. This result underscores the advantage of training regression-based models on cardinality maps, which eliminates kernel-based ambiguities and improves robustness to variations in object size.

\section{Conclusion}
In this paper, we introduced YOLO-Count, a novel open-vocabulary, regression-based object counting model that substantially improves object quantity control in text-to-image generation. By integrating cardinality regression, hybrid strong-weak supervision, and representation alignment, YOLO-Count achieves state-of-the-art counting accuracy, computational efficiency, and robust open-vocabulary generalization. Extensive experiments and ablation studies validate its effectiveness in overcoming the limitations of prior approaches, particularly in handling large object quantities and novel categories. Beyond advancing object counting, YOLO-Count provides a practical and differentiable mechanism for enhancing the controllability of text-to-image models, thereby enabling more precise and reliable multimodal generation and perception.

\paragraph{Acknowledgment} This work is supported by NSF Award IIS-2127544 and NSF Award IIS-2433768. We thank Yuheon Joh for insightful discussions and valuable feedback.

{
    \small
    \bibliographystyle{ieeenat_fullname}
    \bibliography{main}
}

\appendix
\counterwithin{figure}{section}
\counterwithin{table}{section}
\counterwithin{equation}{section}
\setcounter{table}{0}
\setcounter{figure}{0}
\setcounter{equation}{0}
\section{Appendix}

\subsection{Additional Details and Ablations}

\subsubsection{Efficiency Comparison with Other Models}

To assess efficiency, we compare YOLO-Count with other object counting models in terms of backbone architecture, parameter count, and inference speed, as summarized in \cref{model_info}. For inference speed evaluation, we measure the frames per second (FPS) of all models on a single NVIDIA RTX 3090 GPU.

The results demonstrate that YOLO-Count achieves at least a $5\times$ speedup over other high-accuracy models, primarily due to its lightweight YOLO-based backbone, which avoids the computational overhead of heavy transformer-based vision backbones such as GroundingDINO~\cite{liu2024grounding} and CLIP~\cite{radford2021learning}. This combination of high efficiency and strong performance highlights YOLO-Count as a practical, plug-and-play module for integrating accurate quantity control into text-to-image (T2I) generation pipelines.

\begin{table}[htbp]
\centering
\caption{Comparison of model architecture and efficiency.}\label{model_info}
\begin{tabular}{lrrl}
\toprule
Model & \#Params & FPS & Backbone \\
\midrule
CountGD~\cite{amini2025countgd} & 146M & 4.93 & SwinT \\
CLIP-Count~\cite{jiang2023clip} & 101M & 10.34 & ViT \\
VLCounter~\cite{kang2024vlcounter} & 103M & 8.45 & ViT \\
CounTX~\cite{AminiNaieni23} & 97M & 9.98 & SwinT \\
DAVE~\cite{pelhan2024dave} & 150M & 2.37 & SwinT \\
YOLO-Count (\textbf{Ours}) & 68M & 50.41 & CNN \\
\bottomrule
\end{tabular}
\end{table}

\subsubsection{Token Optimization for T2I Control}

\begin{figure}[htb]
    \centering
    \includegraphics[width=\linewidth]{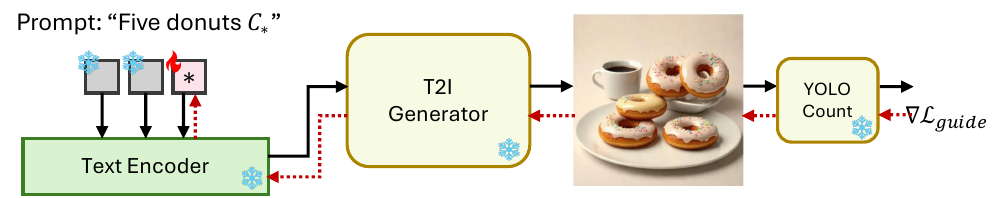}
    \caption{Pipeline for counting-controlled generation via token optimization. ($C_*$ denotes the counting token.)}
    \label{fig:cnt_ctrl}
\end{figure}

We adopt a token optimization strategy analogous to textual inversion. Following \cite{zafar2024iterativeobjectcountoptimization}, we iteratively update the learnable counting token embedding using gradients derived from the discrepancy between the predicted and target counts. This process progressively refines the token representation, guiding the text-to-image (T2I) model to generate images that match the desired object quantity, as illustrated in \cref{fig:cnt_ctrl}. In practice, this optimization requires at most 150 gradient steps and can be executed on a single 32 GB NVIDIA V100 GPU. Optimizing object count for a single image takes approximately 40–180 seconds, depending on the quantity of the target and the complexity of the image.

\subsubsection{Categories for Controllable Generation Benchmarks}

\begin{itemize}
    \item \textbf{LargeGen:} sea shell, apple, orange, marble, green pea, bottle cap, peach, egg, chair, tree log.
    \item \textbf{LargeGen-New:} egg tart, mango, lemon, onion, gold medal, beaker, harmonica, baozi, jellyfish, llama.
\end{itemize}

\begin{figure}[htbp]
    \centering
    \includegraphics[width=\linewidth]{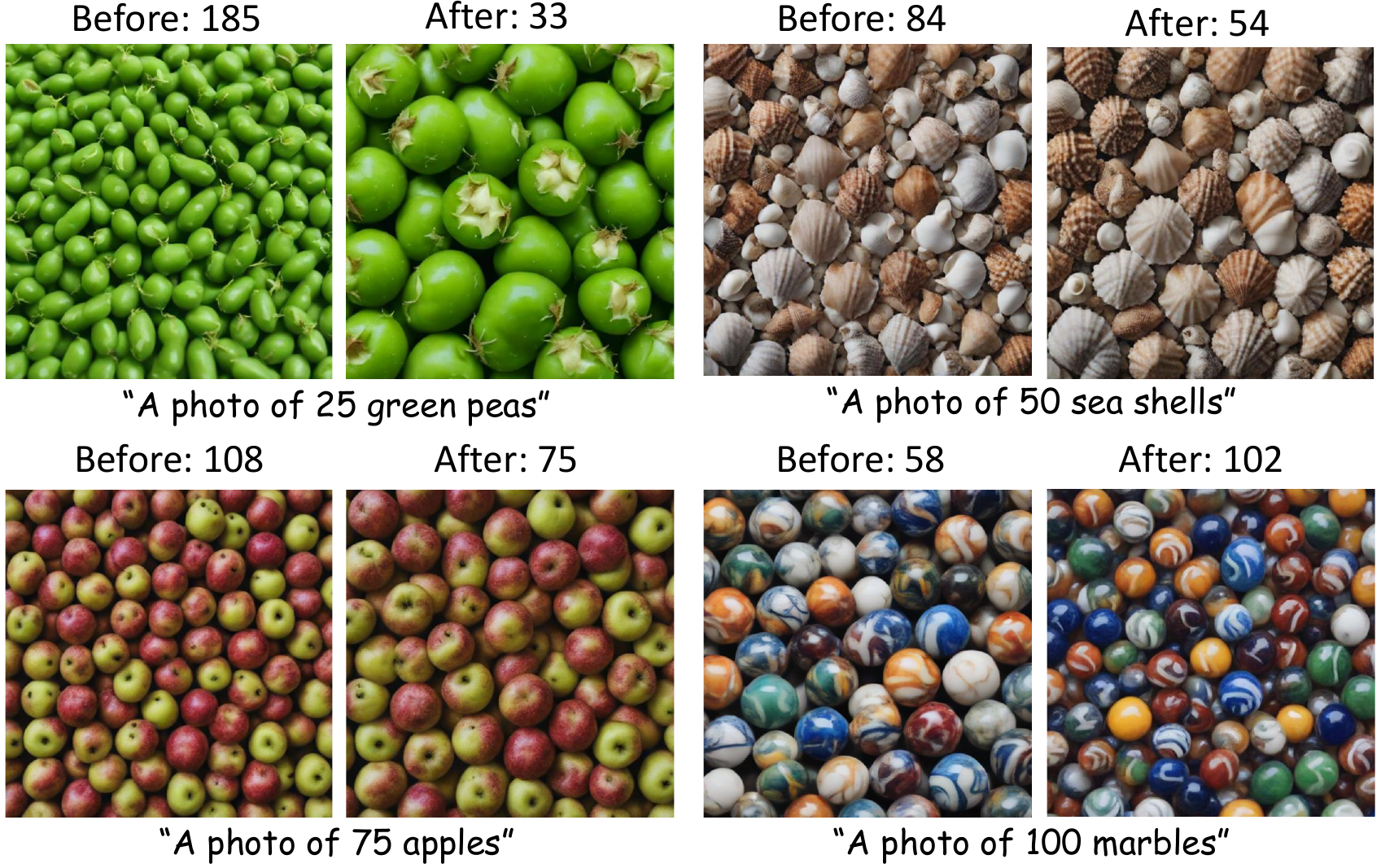}
    \caption{Results of quantity control on LargeGen.}
    \label{fig:additional_largegen}
\end{figure}

\begin{figure}[htbp]
    \centering
    \includegraphics[width=\linewidth]{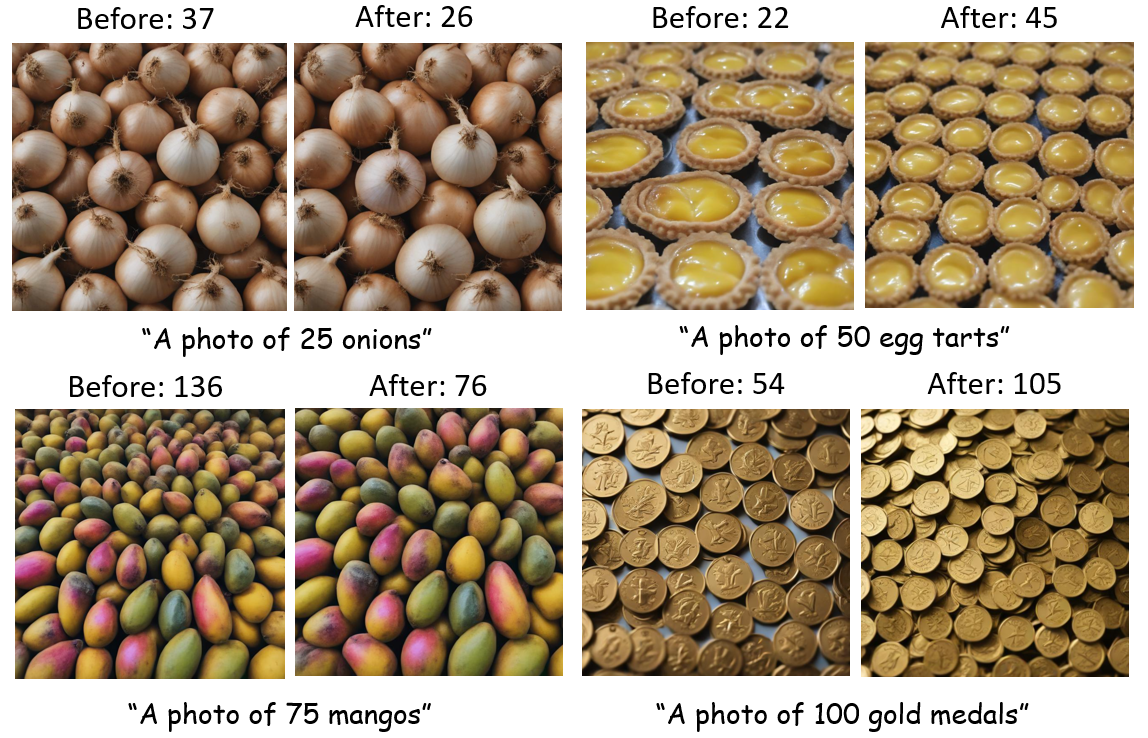}
    \caption{Results of quantity control on LargeGen-New.}
    \label{fig:additional_largegen_new}
\end{figure}

\cref{fig:additional_largegen,fig:additional_largegen_new} shows additional qualitative results of counting-controlled generation on the LargeGen and LargeGen-New benchmarks, respectively, demonstrating YOLO-Count’s ability to accurately guide object quantity control across both seen and novel categories.

\subsubsection{Effect of the Classification Branch During Inference}

The standard inference procedure for YOLO-Count involves summing the cardinality map, as used in all main experiments. However, since YOLO-Count includes an additional classification branch, which is originally designed to aid training, we explore using its classification output to refine inference results.

During inference, we filter the cardinality regression output $\hat{y}_{\mathrm{cnt}}$ using classification probabilities $\hat{y}_{\mathrm{cls}}$. Specifically, only grid cells with classification probabilities exceeding a predefined threshold $\kappa$ are considered valid for counting. The final count is computed as:
\begin{equation}
\mathrm{Count} = \sum_{p \in \mathcal{P}} \hat{y}_\mathrm{cnt}(p),
\end{equation}

where $\mathcal{P} = \{p \mid \hat{y}_\mathrm{cls}(p) > \kappa\}$ represents the set of grid cells whose classification probability surpasses the threshold $\kappa$. We evaluate counting accuracy across different $\kappa$ values on the FSC147~\cite{m_Ranjan-etal-CVPR21} and LVIS~\cite{gupta2019lvis} datasets.

\begin{figure}[htbp]
    \centering
    \includegraphics[width=\linewidth]{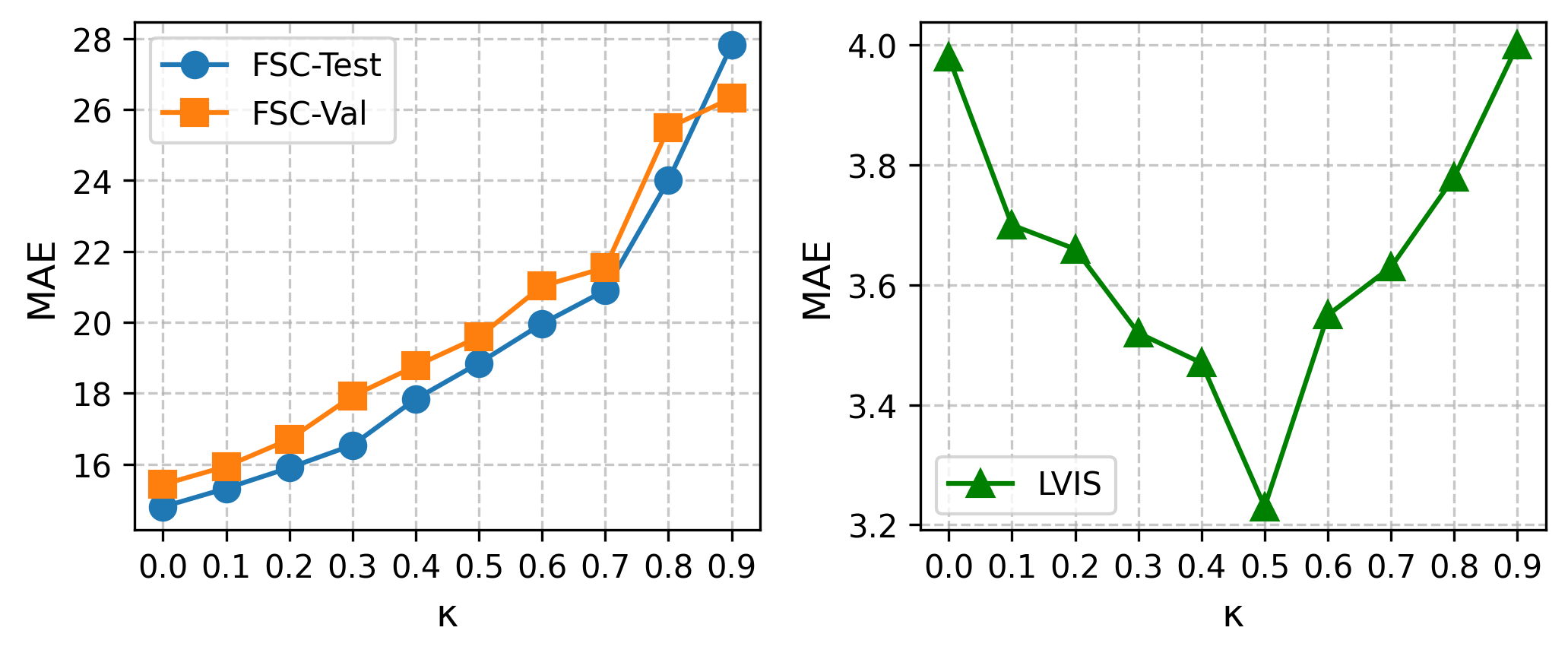}
    \caption{Counting MAE of FSC147 and LVIS under different thresholds.  }
    \label{fig:mae_plot}
\end{figure}

As shown in \cref{fig:mae_plot}, the optimal $\kappa$ differs by dataset: $\kappa=0.0$ achieves the lowest MAE on FSC147, whereas $\kappa=0.5$ performs best on LVIS. This difference reflects dataset-specific annotation protocols. FSC147 employs inclusive labeling, counting any object partially matching the prompt, favoring lower thresholds ($\kappa=0.0$) to avoid missed detections. Conversely, LVIS provides precise multi-category annotations requiring stricter category separation, where moderate thresholds ($\kappa=0.5$) effectively filter visually similar but incorrect categories.

\begin{figure}[ht]
    \centering
    \includegraphics[width=\linewidth]{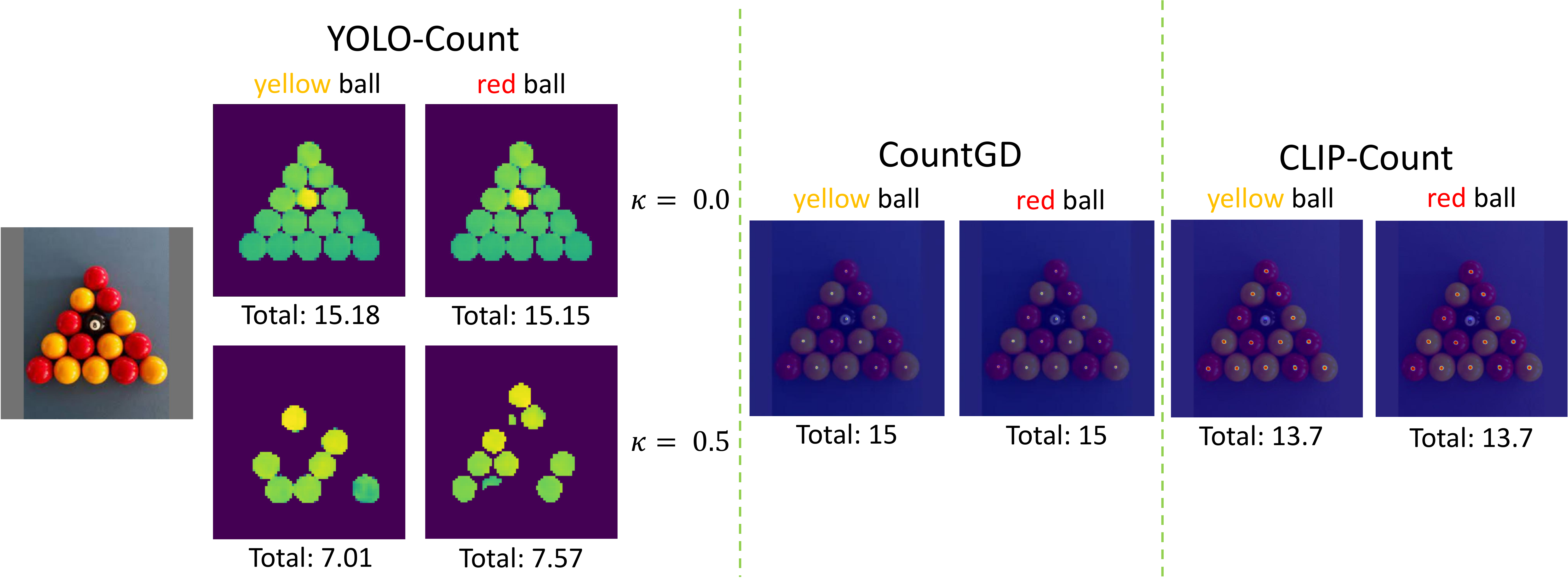}
    \caption{Demonstrations of YOLO-Count in distinguishing semantically similar categories through classification thresholding.}
    \label{fig:threshold}
\end{figure}

\cref{fig:threshold} illustrates this effect in a color-based ball counting task. Baseline models such as CountGD~\cite{amini2025countgd} and CLIP-Count~\cite{jiang2023clip} indiscriminately count all colored balls. In contrast, YOLO-Count adapts based on $\kappa$: at $\kappa=0.0$, it mirrors inclusive counting behavior, while at $\kappa=0.5$, it selectively counts only the target color by leveraging classification filtering. This shows that $\kappa$ serves as an inference-time hyperparameter, allowing flexible adaptation to task requirements, favoring lower thresholds for inclusive counting (FSC147 style) and higher thresholds for strict categorical discrimination (LVIS style).

\subsection{Limitations and Future Work}

\begin{figure}[ht]
\centering
\includegraphics[width=\linewidth]{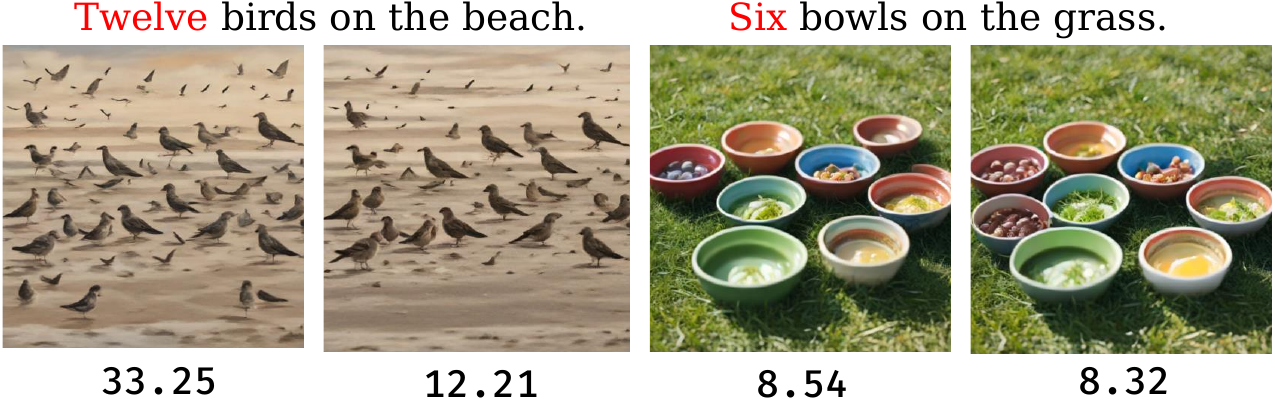}
\caption{Common failure modes. Numbers below are counts.}
\label{fig:failure_case}
\end{figure}

Despite its strong performance, YOLO-Count exhibits several limitations, as illustrated in \cref{fig:failure_case}. On the left, the model suffers from incorrect counting, failing to detect small or background objects such as birds in cluttered scenes. On the right, token optimization fails to reduce the object count toward the specified target, leading to unsuccessful quantity control in challenging scenarios. Furthermore, YOLO-Count’s design is inherently dependent on the YOLO architecture, which, while efficient, restricts its seamless integration with state-of-the-art text-to-image (T2I) diffusion models.

Future work could address these limitations by exploring Transformer-based or hybrid architectures to improve robustness in dense and fine-grained counting scenarios. Additionally, incorporating joint optimization directly within the diffusion process, rather than relying on post-hoc token optimization, may provide stronger and more stable signals for quantity control, enabling tighter coupling between counting models and generative pipelines.

\end{document}